\documentclass[10pt,twocolumn,letterpaper]{article}

\usepackage{iccv}
\usepackage{times}
\usepackage{epsfig}
\usepackage{graphicx}
\usepackage{amsmath}
\usepackage{amssymb}
\usepackage{bm}
\usepackage{balance}
\usepackage{tikz}
\usepackage{array}
\usetikzlibrary{bayesnet}

\usepackage[pagebackref=true,breaklinks=true,letterpaper=true,colorlinks,bookmarks=false]{hyperref}

\newcommand{\N}{\mathcal{N}}
\newcommand{\E}{\mathbb{E}}
\newcommand{\A}{\alpha}

\newcommand{\SN}{\sigma_N}
\newcommand{\SD}{\sigma_D}
\newcommand{\Sp}{\sigma_p}
\newcommand{\G}{\gamma}
\newcommand{\bt}{\bm{\theta}}
\newcommand{\bp}{\bm{\phi}}

\usepackage{color}

\usepackage{pgfplots}

\iccvfinalcopy 

\def\iccvPaperID{2467} 

\ificcvfinal\fi
\begin{document}

\title{\vspace{-5mm}Nonparametric Variational Auto-encoders for \\ Hierarchical Representation Learning}

\author{Prasoon Goyal$^1$~~~~Zhiting Hu$^{1,2}$~~~~Xiaodan Liang$^1$~~~~Chenyu Wang$^2$~~~~Eric P. Xing$^{1,2}$\\
$^1$Carnegie Mellon University~~~ $^2$Petuum Inc.\\
{\tt\footnotesize \{prasoongoyal13,chenyu.wanghao\}@gmail.com, \{zhitingh,xiaodan1,epxing\}@cs.cmu.edu}}

\maketitle

\vspace*{-3mm}
\begin{abstract}
\vspace*{-3mm}
The recently developed variational autoencoders (VAEs) have proved to 
be an effective confluence of the rich representational power
of neural networks with Bayesian methods. However, most work
on VAEs use a rather simple prior over the
latent variables such as standard normal distribution, thereby
restricting its applications to relatively simple 
phenomena. In this work, we propose hierarchical nonparametric  variational autoencoders, which combines tree-structured Bayesian nonparametric priors with VAEs, to enable infinite flexibility of the latent representation space. Both the neural parameters and Bayesian priors are learned jointly using tailored variational inference. The resulting model induces a hierarchical structure of latent semantic concepts underlying the data corpus, and infers accurate representations of data instances. We apply our model in video representation learning. Our method is able to discover highly interpretable activity hierarchies, and obtain improved clustering accuracy and generalization capacity based on the learned rich representations.
\end{abstract}

\thispagestyle{empty}

\vspace{-3mm}
\section{Introduction}
Variational Autoencoders (VAEs)~\cite{kingma2013auto} are among the popular
models for unsupervised representation learning. They consist of a standard 
autoencoder component, that embeds the data into a latent code space by
minimizing reconstruction error, and a Bayesian regularization over the latent space, which enforces the posterior of the hidden code vector matches a prior distribution. 
These models have been successfully applied to 
various representation learning tasks, such as sentence modeling~\cite{bowman2015generating,hu2017controllable} and image understanding~\cite{walker2016uncertain,gregor2015draw}.

However, most of these approaches employ a simple prior over the latent space, which is often the standard normal distribution. Though convenient inference and learning is enabled, converting the data distribution to such fixed, single-mode prior distribution can lead to overly simplified representations which lose rich semantics present in the data.
This is especially true in the context of unsupervised learning where large amount of available data with complex hidden structures is of interest which is unlikely to be presented in the restricted latent space. For example, a large video corpus can encode rich human activity with underlying intricate temporal dependencies and hierarchical relationships. For accurate encoding and new insights into the datasets, it is desirable to develop new representation learning approaches with great modeling flexibility and structured interpretability.

\begin{figure}[t]
\begin{center}
 \includegraphics[width=0.45\linewidth]{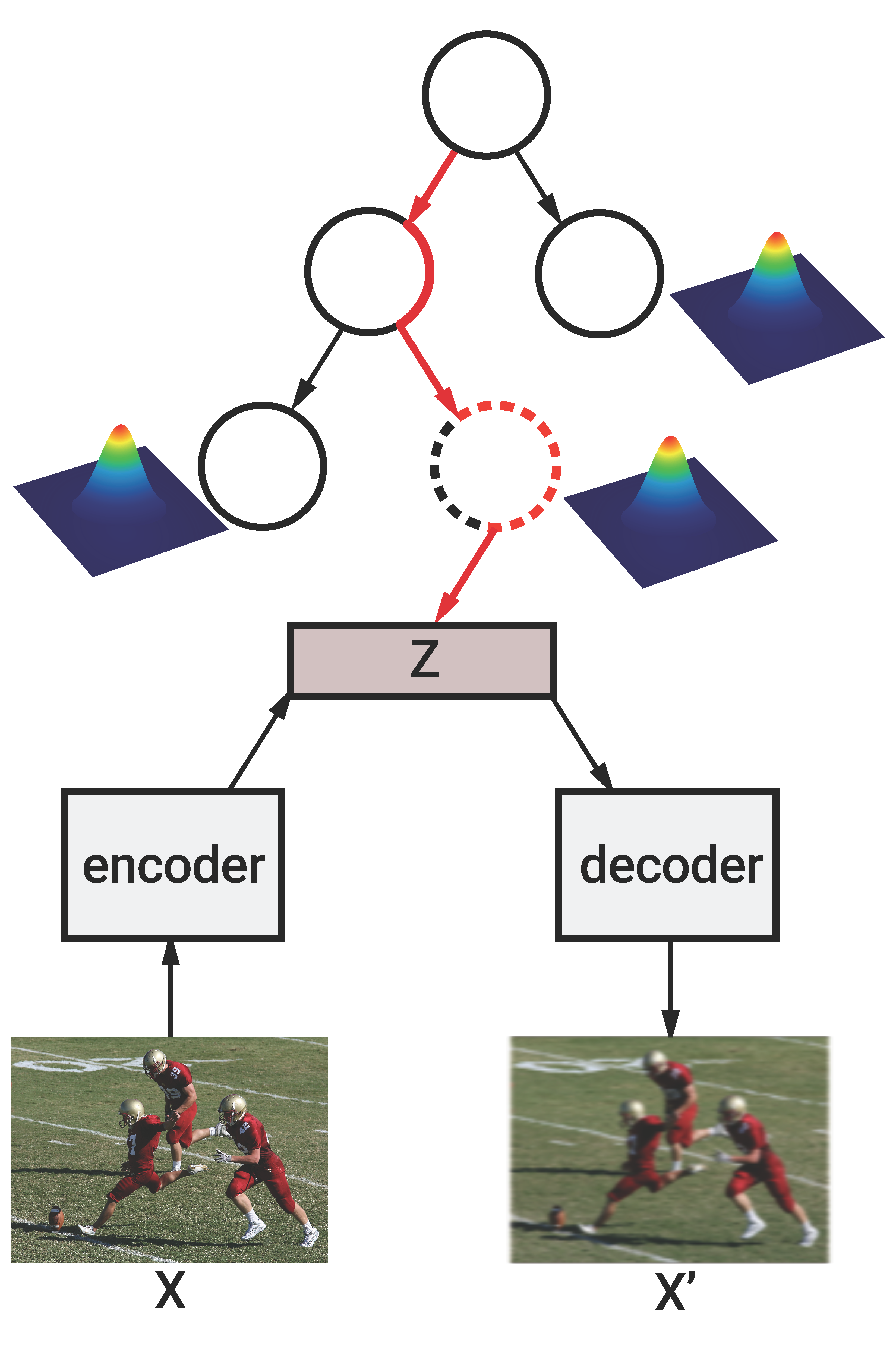}
\end{center}
\vspace{-15pt}
\caption{Illustration of the nonparametric hierarchical variational autoencoder. We combines hierarhical Bayesian nonparametric priors with variational autoencoders.}
\label{fig:overview}
\end{figure}

In this paper, we propose \emph{hierarchical nonparametric  variational autoencoders}, which combines Bayesian nonparametric priors with VAEs. Bayesian nonparametric methods as the code space prior can grow information capacity with the amount and complexity of data, which endows great representational power of the latent code space. In particular, we employ nested Chinese Restaurant Process (nCRP)~\cite{blei2010nested}, a stochastic process allowing infinitely deep and branching trees for representing the data. As opposed to fixed prior distributions in previous work, we learn both the VAE parameters and the nonparametric priors \emph{jointly} from the data, for self-calibrated model capacity. The induced tree hierarchies serve as an aggregated structured representation of the whole corpus, summarizing the gist for convenient navigation and better generalization. On the other hand, each data instance is assigned with a probability distribution over the paths down the trees, from which an \emph{instance-specific} prior distribution is induced for regularizing the instance latent code. Figure~\ref{fig:overview} gives a schematic overview of our approach.

The resulting model unifies the Bayesian nonparametric flexibility with neural inductive biases, by viewing it as a nonparametric topic model~\cite{blei2012probabilistic} on the latent code space, in which raw data examples are first transformed to compact (probabilistic) semantic vectors with deep neural networks (i.e., the encoder networks). This enables invariance to distracting transformations in the raw data~\cite{larsen2015autoencoding,dosovitskiy2016generating}, resulting in robust topical inference.
We derive variational inference updates for estimating all parameters of the neural autoencoder and Bayesian priors jointly. A tailored split-merge process is incorporated for effective exploration of the unbounded tree space.

Our work is the first to combine tree-structured BNPs and VAE neural models in a unified framework, with all parameters learned jointly. From the VAE perspective, we propose the first VAE extension that learns priors of the latent space from data. From the BNP perspective, our model is the first to integrate neural networks for efficient generation and inference in Dirichlet process models.

We present an application on video corpus summarization and representation learning, in which each video is modeled as a mixture of the tree paths. Each frame in the video is embedded to the latent code space and attached to a path sampled from the mixture. The attachment dynamics effectively clusters the videos based on sharing of semantics (e.g., activities present in the video) at multiple level of abstractions, resulting in a hierarchy of abstract-to-concrete activity topics. The induced rich latent representations can enable and improve a variety of downstream applications. We experiment on video classification and retrieval, in which our model obtains superior performance over VAEs with parametric priors. Our method also shows better generalization on test set reconstruction. Qualitative analysis reveals interpretability of the modeling results.

We begin by reviewing related work in \S\ref{sec:related}. We then present our approach in the problem setting of learning hierarchical representations of sequential data (e.g., videos). \S\ref{sec:prelim} describes the problem and provides background on the nCRP prior. \S\ref{sec:model} develops our nonparametric variational autoencoders and derives variational inference for joint estimation of both the neural parameters and Bayesian priors. In \S\ref{sec:video} we apply the model for video representation learning. \S\ref{sec:exp} shows quantitative and qualitative experimental results. We conclude the paper in \S\ref{sec:conclude}.


%

\section{Related Work}\label{sec:related}
\paragraph{Variational autoencoders and variants.}
Variational Autoencoders (VAEs)~\cite{kingma2013auto} provide a powerful framework for deep unsupervised representation learning. VAEs consist of encoder and decoder networks which encode a data example to a latent representation and generate samples from the latent space, respectively. The model is trained by minimizing an expected reconstruction error of observed data under the posterior distribution defined by the encoder network, and at the same time regularizing the posterior of the hidden code to be close to a prior distribution, by minimizing the KL divergence between the two distributions. Vanilla VAEs typically use a standard normal distribution with zero mean and identity covariance matrix as the prior, which enables closed-form optimization while restricting the expressive power of the model. Adversarial autoencoders~\cite{makhzani2015adversarial} replace the KL divergence with an adversarial training criterion to allow richer families of priors. Our work differs in that we compose VAEs with Bayesian nonparametric methods for both flexible prior constraints of individual instances and structured representation induction of the whole corpus. Previous research has combined VAEs with graphical models in different context. Siddharth et al.,~\cite{siddharth2017learning} replace the encoder networks with structured graphical models to enable disentangled semantics of the latent code space. Johnson et al.,~\cite{johnson2016composing} leverage the encoder networks to construct graphical model potentials to avoid feature engineering. Our work is distinct as we aim to combine Bayesian nonparametric flexibility with VAEs, and address the unique inferential complexity involving the hierarchical nonparametric models. Other VAE variants that are orthogonal to our work are proposed. Please refer to \cite{hu2017unifying} for a general discussion of VAEs and their connections to a broad class of deep generative models.


\begin{figure}[t]
\begin{center}
 \includegraphics[width=\linewidth]{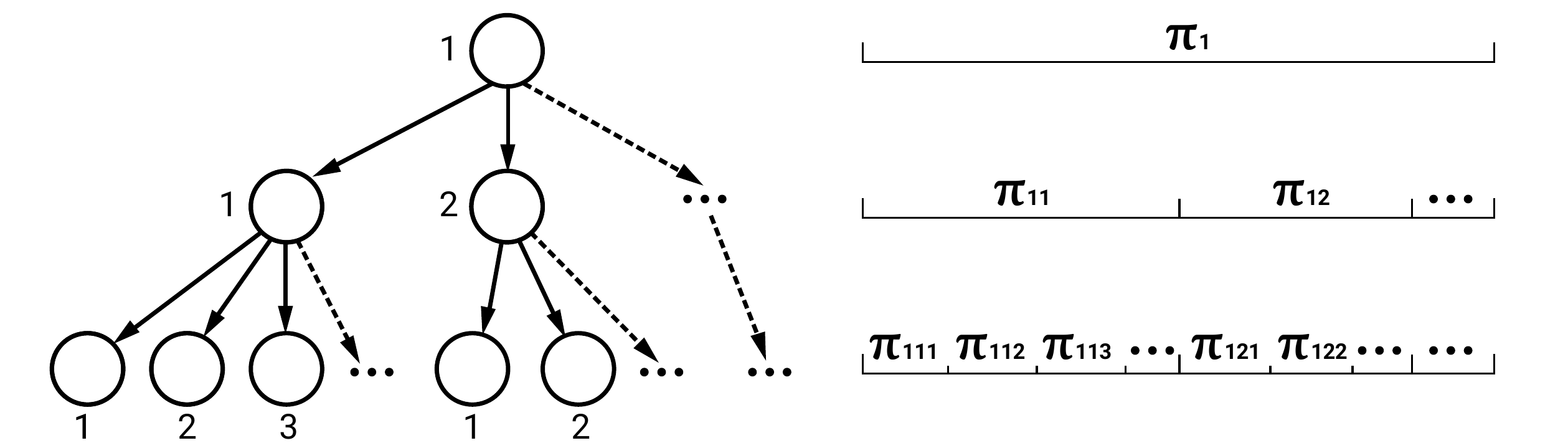}
\end{center}
\vspace{-10pt}
\caption{{\bf Left}: a sample tree structure draw from nCRP. {\bf Right}: The respective tree-based stick-breaking construction. The stick length of the root node is $\pi_1=1$. Each node performs a stick-breaking process on its stick segment to construct its children.}
\label{fig:hdp}
\end{figure}

\paragraph{Bayesian nonparametric methods.}
Bayesian nonparametric methods allow infinite information capacity to capture rich internal structure of data. For example, mixture models with Dirichlet process priors can be used to cluster with an unbounded number of centers. A few recent works have developed powerful hierarchical nonparametric priors~\cite{blei2010nested,hu2015large} to induce tree structures with unbounded width and depth. Nested Chinese Restaurant Process (nCRP) assigns data instances with paths down the trees. The attachment dynamics lead to hierarchical clustering of the data where high-level clusters represent abstract semantics while low-level clusters represent concrete content. We leverage nCRP as the prior over the latent code space for enhanced representational power. Gaussian processes~\cite{rasmussen2006gaussian} are another line of Bayesian nonparametric approach which has been incorporated with deep neural networks for expressive kernel learning~\cite{wilson2016deep,hinton2008using}. These methods have typically been applied in supervised setting, while we are targeting on unsupervised representation learning using hierarchical Dirichlet nonparametrics.



\section{Preliminaries}
\label{sec:prelim}
For concreteness, we present our approach in the problem setting of unsupervised hierarchical representation learning of sequential data.
We start by describing the problem statement, followed
by an overview of nCRP. All the notations used in the
paper have been consolidated in Table~\ref{table:notations}
for quick reference.

\subsection{Problem Description}
\label{sec:problem}
Let $\bm{x}^{m} = (x_{mn})_{n=1}^{N_m}$ denote a sequence $\bm{x}^{m}$ of length $N_{m}$ with $n^{th}$ element denoted as $x_{mn}$. Given unlabeled sequences $\{\bm{x}^{m}\}$ of data, we want to learn compact latent representation for each instance as well as capture the gist of the whole corpus. To this end, we build a generative probabilistic model that assigns high probability to the given data. Further, to capture rich underlying semantic structures, we want the probabilistic model to be hierarchical, that is, coarse-grained concepts are higher up in the hierarchy, and fine-grained concepts form their children.

For instance, video data can be modeled as above, wherein each video can be represented as a sequence $\bm{x}^{m}$. Each element $x_{mn}$ of the sequence is a temporal segment of the video, such as a raw frame or sub-clip of the video, or some latent representation thereof. In such data, the hierarchy should capture high-level activities, such as, ``playing basketball'' higher up in the hierarchy, while more fine-grained activities, such as ``running'' and ``shooting'' should form its children nodes. These hierarchies can then be used for a wide variety of downstream tasks, such as, video retrieval, summarization, and captioning.

\subsection{Nested Chinese Restaurant Process}
\label{sec:hdp}
We use nCRP priors~\cite{blei2010nested}, 
which can be recursively defined in terms of Dirichlet
process (DP). 
A draw from a Dirchlet process $DP(\gamma, G_{0})$ 
is described as
\begin{equation}
\small
\begin{split}
v_{i} \sim Beta(1, \G), \hspace{5mm}
\pi_{i} = v_{i} \prod_{j=1}^{i-1} (1 - v_{j})& \\
w_{i} \sim G_{0}, \hspace{5mm}
 \hspace{7mm}
G = \sum_{i=1}^{\infty} \pi_{i} \delta_{w_i}&
\end{split}
\end{equation}
Here, $\gamma$ is the scaling parameter, $G_{0}$ is
the base distribution of the DP, and $\delta_w$ is an indicator function that takes value $1$ at $w$ and $0$ otherwise. 
The above construction admits
an intuitive stick-breaking interpretation, in which, a unit
length stick is broken at a random location, and $\pi_1$ is 
the length of the resulting left part. The right part is 
broken further, and the length of left part so obtained is 
assigned to $\pi_2$. The process is continued to infinity.
Note that, $\sum_{i=1}^{\infty} \pi_{i} = 1$. Therefore, a draw
from a DP defines a discrete probability distribution
over a countably infinite set.

The above process can be extended to obtain nCRP,
or equivalently, a tree-based stick-breaking process, in which, we start at the root node (level 0), 
and obtain probabilities over its child nodes (level 1) using 
a DP. Then we recursively run a DP on each level 1 node to get probabilities over level 2 nodes, and so on. This defines
a probability distribution over paths of an infinitely wide and infinitely deep tree. Figure~\ref{fig:hdp} gives an illustration of the process.
More formally, we label all the nodes recursively
using a sequence of integers -- the root node has label `1', its children nodes have labels `11', `12', $\ldots$, children nodes of `11' have labels `111', `112', and so on. Now, we can assign probability to every node $p$ based on draws of stick-breaking weights $v$ as follows:
\begin{itemize}
\setlength\itemsep{0.1em}
\item For the root node (level 0),\  $\pi_{1} = 1$
\item For $i^{th}$ node at level 1,\  $\pi_{1i} = \pi_{1} v_{1i} \Pi_{j=1}^{i} (1 - v_{1j})$.
\item For $j^{th}$ child at level 2 of $i^{th}$ level-1 node,
\vspace*{-2mm}
\begin{center}
$\pi_{1ij} = \pi_{1} \pi_{1i} v_{1ij} \Pi_{k=1}^{j} (1 - v_{1ik})$.
\end{center}
\end{itemize}
\vspace*{-2mm}
This process is repeated to infinity. Please refer to~\cite{blei2010nested,wang2009variational} for more details.

\begin{table}[t]
\centering
\footnotesize
\begin{tabular}{r l}
\hline
Symbol & Description      \\ \hline
$(\bm{x})_{N}$    & a sequence of length N, with 
elements $\bm{x_{1}}, \ldots, \bm{x}_{N}$ \\ 
$\bm{x}_{mn}$     & $n^{th}$ element of sequence 
$(\bm{x}^{m})_{N}$ \\
$\bm{z}_{mn}$     & the latent code corresponding
to $\bm{x}_{mn}$ \\
$par(p)$              & the parent node of node $p$ \\
$\bm{\A}_{p}$         & the parameter vector for node 
$p$ \\
$\bm{\A}^{*}$         & the prior parameter over $\A_{p}$ 
for the root node \\
$\SN$                 & the variance parameter for 
node parameters \\
$\SD$                 & the variance parameter for 
data \\
$v_{me}$             & the nCRP variable for $m^{th}$ 
sequence on edge $e$\\
$\bm{V}_{m}$      & the set of all $v_{me}$ for
$m^{th}$ sequence \\
$\G^{*}$              & the prior parameter shared by all 
$v_{me}$ \\
$c_{mn}$              & the path assignment for data point
$\bm{x}_{mn}$ \\
$c_{mn}$              & the path assignment for data point
$\bm{x}_{mn}$ \\
$\bm{\mu}_{p}, \Sp$        & parameters for variational
distribution of $\A_{p}$ \\
$\G_{me,0}, \G_{me,1}$ & parameters for variational
distribution of $v_{me}$ \\
$\bm{\phi}_{mn}$          & parameter for variational
distribution of $c_{mn}$ \\

\hline
\end{tabular}
\vspace{1mm}
\caption{Notations used in the paper}
\label{table:notations}
\end{table}

\section{Hierarchical Nonparametric Variational Autoencoders}
\label{sec:model}

We first give a high-level overview of our framework, and then describe various components in detail.

Formally, a VAE takes in an input $\bm{x}$, 
that is
passed through an encoder with parameters $\bp$ to produce
a distribution $q_{\phi}(\bm{z} 
\vert \bm{x})$ over the
latent space. Then, a latent code $\bm{z}$ is 
sampled from 
this distribution, and passed through a decoder to 
obtain the
reconstructed data point $\tilde{\bm{x}}$. Thus, 
minimizing the
reconstruction error amounts to maximizing 
$\E_{\bm{z} \sim q_{\bp}(\bm{z} \vert 
\bm{x})}
[\log p_{\bt}(\bm{x} \vert 
\bm{z})]$, where 
$p_{\bt}(\bm{x} \vert \bm{z})$ 
corresponds to the 
decoder parameterized by $\bt$.
The encoder and the decoder can be arbitrary 
functions; however, they are typically modeled as 
neural networks.
Further, a prior $p_{\bt}(\bm{z})$ 
is imposed on 
the latent space. Thus we want to solve for
parameters $\bp$ and $\bt$, which,
using the standard variational inference
analysis gives the following lower bound on the
data likelihood:

\vspace*{-5mm}
\begin{equation}
\small
\begin{split}
\log p_{\bt}(\bm{x}^{m}) &\geq  \mathcal{L}(\bt, \bp; \bm{x}^{m}) \\
&= \E_{\bm{z} \sim q_{\bp}(\bm{z} \vert 
\bm{x}^{m})} [\log p_{\bt}(\bm{x}^{m} \vert \bm{z})] \\
&\quad\  - D_{KL}(q_{\bp}(\bm{z} \vert \bm{x}^{m} )\| p_{\bt}  (\bm{z})) \\
\end{split}
\end{equation}
Therefore, the prior and the decoder together act as the generative model of the data, while the encoder network acts as the inference network, mapping data to posterior distributions over the latent space. 
Typically, the prior distribution $p_{\bt}(\bm{z})$ is assumed to be standard normal distribution $\N(0, I)$, 
which implies that maximizing the above lower bound amounts to optimizing only the neural network parameters, since in that case, the prior is free of parameters.

In this work, we use a much richer prior, namely the nCRP prior described in \ref{sec:hdp}. This allows growing information capacity of the latent code space with the amount and complexity of data, and thus obtains accurate latent representations. The tree-based prior also enables automatic discovery of rich semantic structures underlying the data corpus. To this end, we need to jointly optimize for the neural network parameters and the parameters of the nCRP prior. We make use of alternating optimization, wherein we first fix the nCRP parameters and perform several 
backpropagation steps to optimize for the neural network parameters, and then fix the neural network, and perform variational inference updates to optimize for the nCRP parameters. 

We next describe the nCRP-based generative model and variational inference updates.

\begin{figure}
  \begin{center}
%
%
%


\resizebox{8cm}{!}{
\begin{tikzpicture}
\tikzstyle{main}=[circle, minimum size = 10mm, thick, draw =black!80, node distance = 12mm]
\tikzstyle{connect}=[-latex, thick]
\tikzstyle{box}=[rectangle, draw=black!100]
  \node[main, fill = white!100] (alpha) [label=below:$\gamma^*$] { };
  \node[main] (theta) [right=of alpha,label=below:$\mathbf{V_{m}}$] { };
  \node[main] (z) [right=of theta,label=below:$c_{mn}$] {};
  \node[main] (alpha_s) [above=of alpha, label=below:$\bm{\A^*$}] { };
  \node[main] (alpha_p) [right=of alpha_s, label=below:$\bm{\A_{par(p)}}$]
  { };
  \node[main] (beta) [above=of z,
  label=below:$\bm{\A_{p}}$] 
  { };
  \node[main, fill = black!10] (w) [right=of z,label=below:$\mathbf{z_{mn}}$] { };
  \path (alpha) edge [connect] (theta)
        (theta) edge [connect] (z)
		(z) edge [connect] (w)
        (alpha_s) edge [connect] (alpha_p)
        (alpha_p) edge [connect] (beta)
		(beta) edge [connect] (w);
  \node[rectangle, inner sep=0mm, fit= (alpha_p) (beta),label=above right:$K$, xshift=10mm] {};
  \node[rectangle, inner sep=4.4mm, draw=black!100, fit= (alpha_p) (beta)] {};
  \node[rectangle, inner sep=0mm, fit= (z) (w),label=above right:$N_m$, xshift=8mm, yshift=-1mm] 
  {};
  \node[rectangle, inner sep=3.5mm,draw=black!100, fit= (z) (w)] {};
  \node[rectangle, inner sep=0mm, fit= (z) (w),label=above right:M, xshift=13mm, yshift=2mm] {};
  \node[rectangle, inner sep=6.5mm, draw=black!100, fit = (theta) (z) (w)] {};
\end{tikzpicture}
}




  \end{center}
   \vspace{-5pt}
  \caption{The proposed generative model. This diagram only shows the BNP
  component. Thus, the latent codes $z_{mn}$ that are learnt in VAE training
  are treated as observations.}
  \label{fig:plate-notation}
\end{figure}
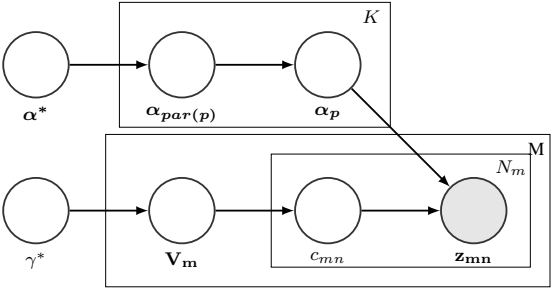

\subsection{Generative Model} \label{sec:genmodel}
The generative model assumes a tree with infinite depth and branches, and generates data sequences through root-to-leaf random walks along the paths of the tree. Each node $p$ has a parameter vector $\bm{\A}_{p}$ which depends on the parameter vector of the parent node to encode the hierarchical relation. That is, for every node $p$ of the tree, draw a $D$-dimensional parameter vector $\bm{\A}_p$, according to 
\vspace*{-2mm}
\begin{equation}
\begin{split}
\bm{\A}_p \sim \N(\bm{\A}_{par(p)}, \SN^2 I)
\end{split}
\vspace*{-3mm}
\end{equation}
where $par(p)$ denotes the parent node of $p$, and $\SN$ is a variance parameter shared by all nodes of the tree. For the root node, we define $\bm{\A}_{par(p)} = \bm{\A}^{*}$, for some constant vector $\bm{\A}^{*}$.

Each data sequence $\bm{x}^m$ is modeled as a mixture of the paths down the tree, and each element $x_{mn}$ is attached to one path sampled from the mixture. Specifically, $\bm{x}^m$ is drawn as follows (Figure \ref{fig:plate-notation} gives the graphical model representation):
\begin{enumerate}
\vspace{-2mm}
\item For each edge $e$ of the tree, draw 
$v_{me} \sim Beta(1, \G^{*})$. We denote the collection of all 
$v_{me}$ for sequence $m$ as $\bm{V}_{m}$.
This defines a distribution over the paths of the tree, as described in section~\ref{sec:hdp}. Let $\pi(\bm{V}_{m})$ denote the probabilities assigned to each leaf node through this process.
%
\vspace*{-2mm}
\item For each element $\bm{x}_{mn}$ in 
$\bm{x}^m$, draw a path $c_{mn}$ according to the multinomial distribution $Mult(\pi(\bm{V}_{m}))$.
\vspace*{-2mm}
\item Draw the latent representation vector $\bm{z}_{mn}$ according to $\N(\bm{\A}_{c_{mn}}, \SD^2 I)$ which is the emission distribution defined by the parameter associated in the leaf node of path $c_{mn}$. Here $\sigma_D$ is a variance parameter shared by all nodes.
%
\end{enumerate}
This process generates a latent code $\bm{z}_{mn}$, 
which is then passed through the decoder to get the observed data $\bm{x}_{mn}$. To summarize the above generative process, the node parameters of the tree depend on their parent node, and the tree is shared by the entire corpus. For each sequence, draws $\bm{V}_{m}$ define a  distribution over the paths of the tree. For each element of the sequence, a path is sampled according to the above distribution, and finally, the data element is drawn according to the node parameter of the sampled path.


Our goal is to estimate the parameters of the tree model, including the node parameters $\bm{\A}_{p}$, sequence-level parameters $\bm{V}_{m}$, and path assignments $c$, as well as the neural parameters $\bt$ and $\bp$, given the hyperparameters $\{\bm{\A}^{*}, \G^{*}, \SN, \SD\}$ and the data.

\subsection{Parameter Learning}
\label{sec:learning}
In this section, we first describe the variational inference updates for estimating the parameters of nCRP prior
(section~\ref{sec:varinf}), and the update equations for our neural network parameters (section~\ref{sec:nn-updates}). Finally, we describe a procedure for joint optimization of the nCRP prior parameters and the neural network.

\subsubsection{Variational Inference}
\label{sec:varinf}
Using the mean-field approximation, we assume the following
forms of the variational distributions:

\begin{itemize}
\setlength\itemsep{0.1em}
\item For each node $p$ of the tree, the parameter vector
$\bm{\A}_p$ is distributed as $\bm{\A}_{p} \sim 
\N(\bm{\mu}_{p}, \Sp^2 I)$, where $\bm{\mu}_{p}$ is a $D$-dimensional 
vector and $\Sp$ is a scalar.
\item For sequence $m$, the DP variable at edge $e$, $v_{me}$ is
distributed as $v_{me} \sim 
Beta(\G_{me, 0}, \G_{me, 1})$, where $\G_{me, 0}$
and $\G_{me, 1}$ are scalars.
\item For data $\bm{x}_{mn}$, the path assignment variable $c_{mn}$ is
distributed as $c_{mn} \sim Mult(\bm{\phi}_{mn})$, 
where the dimension of $\bm{\phi}_{mn}$ is equal to the 
number of paths in the tree.
\end{itemize}


%

We want to find optimal variational parameters that maximize the variational lower bound
\begin{equation}
\small
\mathcal{L} = \E_{q}[\log p(W, X \vert \Theta)] - 
\E_{q}[\log q_{\nu}(W)]
\end{equation}
where $W$ denotes the collection of latent 
variables, $X = \{z_{mn}\}$ are the latent vector representations of observations, $\Theta$ are the hyperparameters, and $\nu = \{\bm{\mu}_{p}, \Sp, \G_{me, 0}, \G_{me, 1}, \bm{\phi}_{mn}\}$ are variational parameters. We use $p(W, X \vert \Theta)$ to denote the generative model described in section \ref{sec:genmodel}. 
We derive variational inference for a {\it truncated} tree~\cite{hu2015large,wang2009variational}. We achieve this by setting the components corresponding to all other paths of $\bm{\phi}_{mn}$ equal to $0$ for all $m \in \{1, \ldots, M\}$ 
and $n \in \{1, \ldots, N_{m}\}$. We later describe how we
can dynamically grow and prune the tree during training.
Thus, the generative distribution above simplifies to the
following:
\begin{align}
\small
&p(W, X \vert \Theta) \\
&= \sum_{p} \log p(\bm{\A}_{p} \vert \bm{\A}_{par(p)}, 
\SN) \nonumber + \sum_{m,e} \log p(v_{me} \vert 
\G^*) \nonumber \\
&+ \sum_{m,n} \log p(c_{mn} \vert 
\bm{V_{m}}) \nonumber + \log p(\bm{z}_{mn} 
\vert \bm{\A}, c_{mn}, \SD)
\end{align} 
Here, $p \in \{1, \ldots, P\}$ and $e \in \{1, \ldots, E\}$ index the 
paths and the edges of the truncated tree respectively.
Note that the above truncation is nested. That is, for two
trees $T_{1}$ and $T_{2}$ such that the set of nodes of $T_{1}$
is a subset of the set of nodes of $T_{2}$, the model
generated from $T_2$ subsumes all possible configurations
that can be generated from $T_1$.

Proceeding as in standard derivation of posterior estimate, we 
obtain the following variational updates:
\vspace{-2mm}
\begin{align}
\small
q^*(\bm{\A}_p \vert \bm{\mu}_p, \Sp)
\sim \N(\bm{\mu}_{p}, \Sp^2)
\end{align}
where, for a leaf node,
\begin{align}
\small
\frac{1}{\Sp^2} &= \frac{1}{\SN^2} + 
\frac{\sum_{m=1}^{M}\sum_{n=1}^{N_{m}}\phi_{mnp}}{\SD^2} \\
\vspace{-2mm}
\bm{\mu}_{p} &= \sigma_{p}^2 \cdot \left( 
\frac{\bm{\mu}_{par(p)}}{\SN^2} + 
\frac{\sum_{m=1}^{M}\sum_{n=1}^{N_{m}}
\phi_{mnp}\bm{z}_{mn}}{\SD^2}
\right)
\end{align}
while for an internal node:
\vspace{-2mm}
\begin{align}
\small
\frac{1}{\Sp^2} &= \frac{1 + \vert ch(p) \vert}
{\SN^2} \\
\mu_{p} &= \Sp^2 \cdot \left( 
\frac{\bm{\mu}_{par(p)} + \sum_{r \in ch(p)} 
\bm{\mu}_{r}}{\SN^2} 
\right)
\end{align}
Here, $ch(p)$ denotes the set of all children of node $p$, and
$\vert \cdot \vert$ denotes the cardinality of a set.
Intuitively, for a leaf node, $\Sp$ is small when we have
many points (high $\phi_{mnp}$) associated with this node,
which corresponds to a good estimate of parameter 
$\bm{\A_{p}}$.
The mean for a leaf node, $\bm{\mu}_{p}$, 
is a weighted mean of the latent codes of the data. 
For an internal node, the mean parameter, $\bm{\mu}_{p}$ is a
simple average of all the child nodes (and the parent node).
However, a child node with larger amount of data is 
farther from its parent node, and thereby has a greater 
effect on the mean implicitly.
\begin{align}
\small
q^*(v_{me} \vert \G_{me, 0}, \G_{me, 1})
\sim Beta(\G_{me, 0}, \G_{me, 1})
\end{align}
\vspace{-2mm}
where
\vspace{-5mm}
\begin{align}
\small
\G_{me, 0} &= 1 + \sum_{n=1}^{N_{m}} \sum_{p : e \in p} \phi_{mnp} \\
\G_{me, 1} &= \G^* + 
\sum_{n=1}^{N_{m}} \sum_{p : e < p} \phi_{mnp}
\end{align}
Here, $e \in p$ denotes the set of all edges that lie on path $p$,
while $e < p$ denotes the set of all edges that lie to the left
of $p$ in the tree.
\begin{align}
\small
q^*(c_{mn} \vert \bm{\phi}_{mn})
&\sim Mult(\bm{\phi}_{mn})
\end{align}
\vspace{-2mm}
where
\begin{align}
\small
\phi_{mnp} 
\propto 
\exp \bigg\{ 
&\sum_{e : e \in p} [\Psi(\G_{me, 0}) - \Psi(\G_{me, 0} 
+ \G_{me, 1})] \nonumber \\
+ & \sum_{e : e < p} [\Psi(\G_{me, 1}) - \Psi(\G_{me, 0} +
\G_{me, 1})] \nonumber \\
- & \frac{1}{2\SD^2} \left[
(\bm{z}_{mn} - \bm{\mu}_{p})^T 
(\bm{z}_{mn} - \bm{\mu}_{p}) + \sigma_{p}^2 \right]
\bigg\}
\end{align}
Here, $\Psi(\cdot)$ is the digamma function.

\subsubsection{Neural Network Parameter Updates}
\label{sec:nn-updates}

The goal of neural network training is to maximize the following
lower bound on the data log-likelihood function:
\begin{align}
\mathcal{L} = \E_{\bm{z} \sim q_{\bp}(\bm{z} \vert 
\bm{x}^{m})} [\log p_{\bt}(\bm{x} \vert \bm{z})]
- D_{KL}(q_{\bp}(\bm{z} \vert \bm{x} )\| p_{\bt}  (\bm{z}))
\end{align}
with respect to the neural network parameters.
Note that $\bp$ denotes the parameters of the encoder network,
while $\bt$ denotes the parameters of the decoder network and
the nCRP prior. Defining $\bt_{NN}$ as the set of parameters
of the decoder network, we need to learn parameters
$\{ \bp, \bt_{NN}\}$.

The update equations for a parameter $\beta \in \{ \bp, \bt_{NN}\}$
is given by
\begin{equation}
\small
\beta^{(t+1)} \leftarrow \beta^{(t)} + \eta \cdot
\dfrac{\partial \mathcal{L}}{\partial \beta}
\end{equation}
where the partial derivative is computed using backpropagation
algorithm, while $\eta$ is an appropriate learning rate.

\subsubsection{Joint Training}\label{sec:joint}
In order to jointly learn the nCRP parameters and 
the NN parameters, we employ alternating optimization, 
wherein, we first fix the nCRP prior parameters and
perform several steps of NN parameter updates, and then
fix the NN parameters and perform several steps of nCRP parameter
updates. This enables the variational inference to use increasingly
accurate latent codes to build the hierarchy, and the
continuously improving hierarchy guides the neural network to learn more semantically meaningful latent codes.

\subsection{Dynamically Adapting the Tree Structure}
Since our generative model is non-parametric, it admits growing
or pruning the tree dynamically, depending on the richness of
the data. Here, we list the heuristics we use for dynamically
growing and pruning the tree.
Note that each data point $\bm{x}_{mn}$ has soft assignments to 
paths, given by $\bm{\phi}_{mn}$. We use these soft assignments
to make decisions about dynamically adapting the tree structure.

\paragraph{Growing the tree} 
We define \emph{weighted radius} of leaf node $p$ as
\begin{equation}
\small
r_{p} = \sqrt{
\dfrac{\sum_{m=1}^{M}\sum_{n=1}^{N_m} 
\phi_{mnp} (\bm{z}_{mn} - \bm{\mu}_{p})^T 
(\bm{z}_{mn} - \bm{\mu}_{p})}
{\sum_{m=1}^{M}\sum_{n=1}^{N_m} \phi_{mnp}}}
\end{equation}
If the weighted radius $r_{p}$ is greater than a threshold $R$,
then we split the leaf node into $K$ children nodes.

\paragraph{Pruning the tree}
For a leaf node $p$, we can compute the total fraction of the data
assigned to this node as
\begin{equation}
\small
f_{p} = \dfrac{\sum_{m=1}^{M}\sum_{n=1}^{N_m} \phi_{mnp}}
{\sum_{m=1}^{M} N_m}
\end{equation}
If the data fraction $f_{p}$ is less than a threshold $F$, then
the leaf node is eliminated. 
If an internal node is left with only one child, then it
is replaced by the child node, thus effectively eliminating
the internal node.
The parameters $R$ and $K$ for growing the tree, and the parameter
$F$ for pruning the tree are set using the validation
set.

\section{Video Hierarchical Representation Learning}\label{sec:video}

In this section, we describe how we can apply our
proposed model to learn meaningful hierarchical 
representations for video data.

Consider an unlabeled set of videos. We want to build a hierarchy in which the leaf nodes represent fine-grained activities, while as we
move up the hierarchy, we obtain more coarse-grained
activities. For instance, a node in the hierarchy may
represent ``sports'', its child nodes may represent
specific sports, such as ``basketball'' and ``swimming''.
The node ``swimming'' can, in turn, have child nodes
representing ``diving'', ``backstroke'', etc.

To use the above framework, we treat each video as a
discrete sequence of frames, by sampling frames from
the video. Then, each frame is passed through a 
pre-trained convolutional neural network (CNN) to obtain
frame features. The resulting frame features are then
used as sequence elements $\bm{x}_{mn}$ 
in our framework. Note, however, that our framework is sufficiently general, and therefore, instead of using the frame features extracted from a pre-trained CNN, we can use the raw frames directly, or even model the video as a discrete sequence of subshots, instead of a discrete sequence of frames.

We optimize the neural and nCRP parameters jointly as described in section~\ref{sec:learning}. This process gives us a posterior estimate of the
nCRP parameters, which we can use to build a hierarchy
for the corpus, as follows. 
We obtain a distribution $\N(\bm{\mu}_{p},
\Sp)$ for each node parameter $\bm{\A}_{p}$. Thus, we can
pass a frame feature vector $\bm{x}$
through the trained encoder
network to get a latent code $\bm{z}$, 
and then assign the frame
to the path whose $\bm{\A}_p$ is closest to $\bm{z}$.
Doing this for all frames results in each node
being associated with the most representative frames
of the activity the node represents.

\section{Experiments}\label{sec:exp}
Here, we present quantitative and qualitative analysis of our
proposed framework. It is worth pointing out that because of the
unavailability of data labeled both with coarse-grained and
fine-grained activities, we conduct quantitative
analysis on the video classification task and video retrieval task,
and qualitatively show the interpretable hierarchy generated by our non-parametric model. 


\subsection{Experimental Settings}
\paragraph{Dataset}
We evaluate the models on TRECVID Multimedia Event Detection (MED) 2011 dataset
\cite{over2014trecvid}, which consists
of 9746 videos in total. Each video is labeled with one of 15 event classes or supplied as a background video without any assigned action label.
In our experiments, we used only the labeled videos of MED dataset, where 1241 videos are used for training, 138 for validation and 1169 for testing.  The mean length of these videos is about 3 minutes.

\paragraph{Feature extraction}
For each video, we extract one frame for every five seconds, resulting in 42393 frames for training, 5218
frames for validation, and 41144 frames for evaluation.
Then, each frame is passed through a VGG-16 network
\cite{simonyan2014very} trained on ImageNet dataset. The output
of the first fully-connected layer is used as the 
4096-dimensional feature vector.

\begin{figure}
  \begin{center}
%
%
%


\resizebox{8cm}{5cm}{
\begin{tikzpicture}[x=1.7cm,y=1.8cm]

\pgfmathdeclarefunction{gauss}{2}{%
  \pgfmathparse{1/(#2*sqrt(2*pi))*exp(-((x-#1)^2)/(2*#2^2))}%
}

\tikzstyle{main}=[circle, minimum size = 10mm, thick, draw =black!80, node distance = 12mm]
\tikzstyle{connect}=[-latex, thick]
\tikzstyle{box}=[rectangle, draw=black!100]
  \node[rectangle, draw=black!100, minimum width = 2mm, minimum height = 30mm] (input) [label=below:$\textbf{x}$] {};
  \node[rectangle, right = of input, minimum width = 2mm, minimum height = 1mm] (inv1) [label=] {};
  \node[rectangle, above =0.5cm of inv1, draw=black!100, minimum width = 2mm, minimum height = 20mm] (zm) [label=below:$\textbf{z}_{mean}$] {};
  \node[rectangle, below =0.5cm of inv1, draw=black!100, minimum width = 2mm, minimum height = 20mm] (zs) [label=below:$\textbf{z}_{stdev}$] {};
  \node[circle, right = of inv1, draw=black!100, minimum size = 10mm] 
  (sample) [label=] {};
  \node[rectangle, right = of sample, draw=black!100, minimum width = 2mm, minimum height = 20mm] (z) [label=below:$\textbf{z}$] {};
  \node[rectangle, right = of z, draw=black!100, minimum width = 2mm, minimum height = 30mm] (xrec) [label=below:$\textbf{x}_{rec}$] {};
  
\begin{axis}[style={samples=200,smooth},
    axis lines=none, anchor=origin, at={(2.9cm,-0.1cm)},x=0.1cm,y=1cm]
\addplot[mark=none] {gauss(0,1)};
\end{axis}

\path (input) edge [connect] (zm)
        (input) edge [connect] (zs)
        (zm) edge [connect] (sample)
        (zs) edge [connect] (sample)
        (sample) edge [connect] (z)
        (z) edge [connect] (xrec);

\end{tikzpicture}
}




  \end{center}
  \vspace{-10pt}
  \caption{The neural network architecture. The input \textbf{x} is a 
  4096-dim VGG feature vector, that is mapped to 48-dim
  vectors $\textbf{z}_{mean}$ and $\textbf{z}_{stdev}$ using one fully 
  connected layer each. A latent code \textbf{z} is
  then sampled from a Gaussian distribution defined by
  $\textbf{z}_{mean}$ and $\textbf{z}_{stdev}$, which is decoded to $\textbf{x}_{rec}$ 
  using one fully connected layer.}
  \label{fig:nn-arch}
\end{figure}
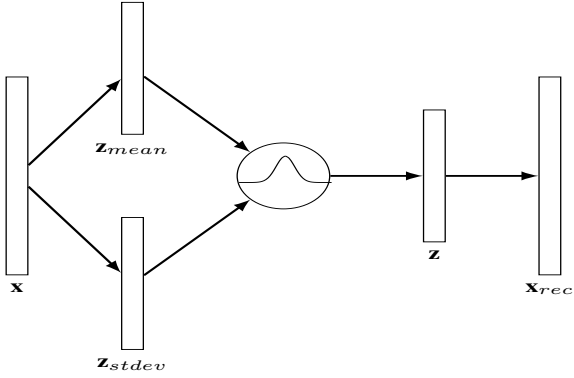

\vspace{-4mm}
\paragraph{Neural network architecture}
Both the encoder and the decoder networks were multi-layer 
perceptions (MLPs). The detailed network is shown in Fig \ref{fig:nn-arch}.
In the alternating optimization procedure we performed one iteration of variational inference updates after every epoch of neural network training. We used RMSProp optimizer~\cite{tieleman2012lecture} with an initial learning rate of $0.01$ and a decay rate of $0.98$ per 1000 iterations. Our model converged in about 20 epochs.

%

\subsection{Test Set Reconstruction}
To better demonstrate the effectiveness of learning hierarchical prior distribution by our non-parametric VAE, we compare the
test-set log likelihood of conventional VAE
with our model. Formally, the log likelihood of
data point $\bm{x}$ is
given by
$\E_{\bm{z} \sim q_{\bm{\phi}}(\bm{z} \vert 
\bm{x})}
[\log p_{\bm{\theta}}(\bm{x} \vert 
\bm{z})]$,
where $q_{\bm{\phi}}(\bm{z} \vert 
\bm{x})$ corresponds to the encoder network and $p_{\bm{\theta}}(\bm{x} \vert 
\bm{z})$ indicates the decoder network.

\begin{table}[t]
\small
\centering\renewcommand\arraystretch{1.5}
\begin{tabular}{l l}
\hline
\textbf{Algorithm\qquad\qquad\quad} & \textbf{\begin{tabular}[c]{@{}c@{}}Mean test log-likelihood\end{tabular}} \\ \hline
VAE-StdNormal      & -28886.90                                                                             \\ \hline
VAE-nCRP            & \textbf{-28438.32}                                                                    \\ \hline
\end{tabular}
\vspace{1mm}
\caption{Test-set log-likelihoods by our model ``VAE-nCRP'' and traditional variational autoencoder ``VAE-StdNormal''.}
\label{table:loglik}
\end{table}

We computed the average log likelihood of the
test set across 3 independent runs of 
variational autoencoders with standard normal
prior, and with nCRP prior. We report the sum of log likelihood
over all frames in the test set.
The results are
summarized in Table \ref{table:loglik}. Our
model obtains a higher log likelihood, implying
that it can better model the underlying complex data distribution embedded in natural diverse videos.
This supports our claim that richer prior distributions
are beneficial for capturing the rich semantics embedded in the data, especially for complex video content.

\begin{table}[t]
\centering\renewcommand\arraystretch{1.2}\setlength{\tabcolsep}{2pt}
\setlength\extrarowheight{-1pt}
\small
\begin{tabular}{l l l l}
\hline
\textbf{Category}            & \textbf{K-Means} & \textbf{VAE-GMM}  & \textbf{VAE-nCRP} \\
\hline
Board\_trick                 & 44.6             & \textbf{47.2} & 31.3          \\
Feeding\_an\_animal          & \textbf{57.0}    & 42.5          & 53.8          \\
Fishing                      & 33.7             & 39.0          & \textbf{48.9} \\
Woodworking                  & 38.9             & 40.5          & \textbf{60.8} \\
Wedding\_ceremony            & 59.8             & 54.3          & \textbf{63.6} \\
Birthday\_party              & 6.5              & 7.4           & \textbf{27.8} \\
Changing\_a\_vehicle\_tire   & 31.9             & 39.7          & \textbf{45.3} \\
Flash\_mob\_gathering        & \textbf{43.4}    & 40.1          & 38.2          \\
Getting\_a\_vehicle\_unstuck & 52.9             & 50.6          & \textbf{65.9} \\
Grooming\_an\_animal         & 2.9              & 14.5          & \textbf{17.3} \\
Making\_a\_sandwich          & 47.1             & \textbf{54.7} & 49.3          \\
Parade                       & 28.4             & \textbf{33.8} & 19.8          \\
Parkour                      & 4.5              & 19.8          & \textbf{27.7} \\
Repairing\_an\_appliance     & 42.3             & \textbf{58.6} & 47.4          \\
Sewing\_project              & 1.6              & \textbf{24.3} & 18.4          \\
\hline
Aggregate over all classes   & 34.9             & 39.1          & \textbf{42.4} \\
\hline

\end{tabular}
\vspace{1mm}
\caption{Classification Accuracy (\%) on TRECVID MED 2011.}
\label{table:accuracy}
\end{table}

\subsection{Video Classification}
We compared our model (denoted as \texttt{VAE-nCRP}) 
with two clustering baselines,
namely, K-Means clustering (denoted as 
\texttt{K-Means}) and variational 
autoencoders with Gaussian mixture model prior
(denoted as \texttt{VAE-GMM}).

\begin{figure*}[!t]
  	\begin{center}
  		\includegraphics[scale = 0.63]{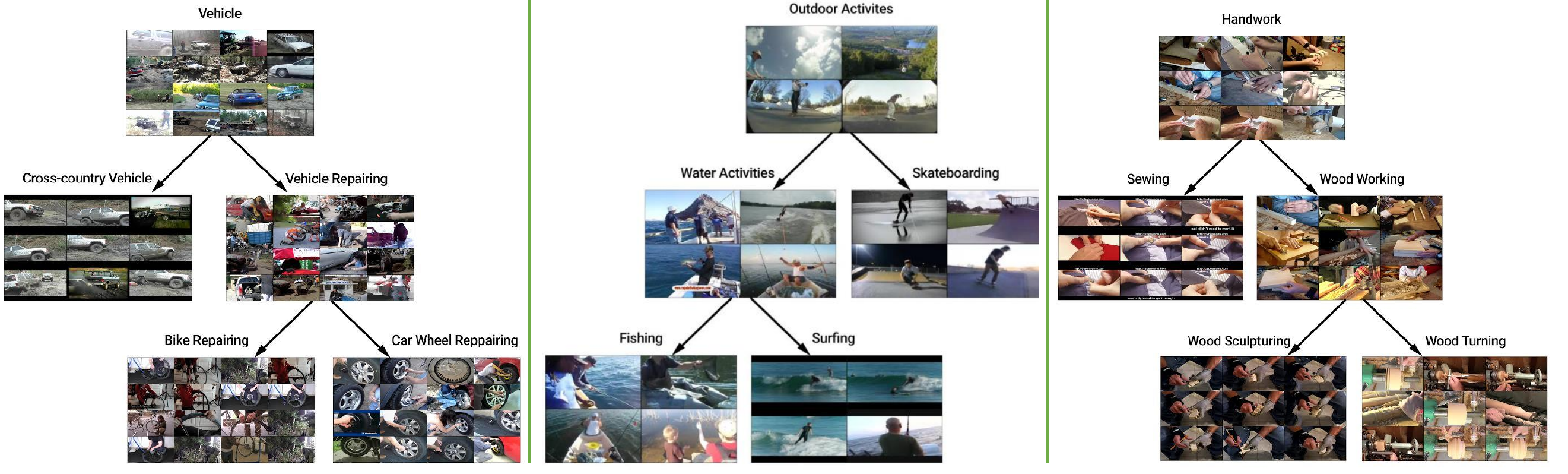}
  		\caption{{Some example hierarchical
structures learned by our model.}}
  		\label{fig:comparison}
  		\vspace{-5mm}
  	\end{center}
      \end{figure*}

In order to evaluate the quality of the obtained hierarchy for the data, our model learns to 
assign an action label to each node (either leaf nodes or internal
nodes) by taking a majority
vote of the labels assigned to the data points. For each frame in the test data, 
we can obtain the latent representation
of the frame feature, and then find the leaf node to
which it is assigned by minimizing the Euclidean
distance between
the latent representation and the leaf node parameter
$\bm{\A}_{p}$, and the predicted label of this frame is then
given by the label assigned to this leaf node. The classification accuracy is then a measure of the
quality of the hierarchy. Similarly, for other clustering baselines,
we assign a label to each cluster, and then assign new data points
to the closest cluster to predict their labels.

Note that, in applications, we would typically
use the standard variational inference framework to find the path assignments, in which case, an unseen frame can also be assigned to a new path, exploiting the non-parametric nature of our model. However, for the purpose of our evaluations, we need to assign a 
label to each frame, and therefore, it must be assigned to one of the paths created during training. We would also like to point out that the hierarchy is constructed in a purely unsupervised manner, and the class labels are used only for evaluation.

We report the mean accuracy of each model, averaged
over three independent runs.
The results are summarized in Table \ref{table:accuracy}.
As can be seen, VAE-nCRP outperforms the 
baseline models on 8 out of 15 classes, and also has an overall
highest accuracy. This suggests that the clusters
formed by VAE-nCRP are more separated than those formed
by K-means clustering and VAE-GMM.

\vspace{-2mm}
\subsection{Video Retrieval}
We also conduct experiments on video retrieval task to further verify the capability of our model. This task aims to retrieve all frames from the test set that belong to each class, which is closely related to video classification task. We report the F-1 scores of the models, which incorporates both the  false positive rate and false negative rate. 
The results are summarized in Table \ref{table:retrieval}.
Again, it can be observed that VAE-nCRP outperforms the baseline models on 8 out of 15 classes, and achieves the highest overall
F-1 score. 

\begin{table}[h]
\centering\renewcommand\arraystretch{1.2}\setlength{\tabcolsep}{2pt}
\setlength\extrarowheight{-1pt}
\small
\begin{tabular}{l l l l}
\hline
\textbf{Category}            & \textbf{K-Means} & \textbf{VAE-GMM}  & \textbf{VAE-nCRP} \\
\hline
Board\_trick                 & 32.1             & \textbf{38.9}          & 32.1          \\
Feeding\_an\_animal          & 33.7             & 33.8          & \textbf{36.2}          \\
Fishing                      & 44.9             & 45.9          & \textbf{59.9} \\
Woodworking                  & 32.1             & 29.5          & \textbf{38.0} \\
Wedding\_ceremony            & 41.0             & \textbf{51.2}          & 51.0 \\
Birthday\_party              & 14.0             & 11.0          & \textbf{30.5} \\
Changing\_a\_vehicle\_tire   & 38.3             & 45.5          & \textbf{54.5} \\
Flash\_mob\_gathering        & \textbf{45.8}             & 41.6          & 42.3          \\
Getting\_a\_vehicle\_unstuck & 37.9             & 43.2          & \textbf{56.9} \\
Grooming\_an\_animal         & 6.7              & \textbf{21.5}          & 20.1 \\
Making\_a\_sandwich          & 51.7             & 53.0          & \textbf{60.3}          \\
Parade                       & 24.7             & \textbf{37.7}          & 29.8          \\
Parkour                      & 6.8              & 28.2          & \textbf{39.1} \\
Repairing\_an\_appliance     & 39.9             & \textbf{41.2}          & 36.8          \\
Sewing\_project              & 1.7              & \textbf{32.5}          & 25.8          \\
\hline
Aggregate over all classes   & 32.4             & 38.5          & \textbf{42.4} \\
\hline

\end{tabular}
\vspace{1mm}
\caption{F-1 scores of video retrieval on TRECVID MED 2011.}
\vspace{-4mm}
\label{table:retrieval}
\end{table}

\vspace{-2mm}
\subsection{Qualitative Analysis}

In addition to the quantitative analysis, we also
performed a qualitative analysis of the hierarchy learned by our model as shown in Figure \ref{fig:comparison}. We visualized the hierarchy structure by representing each node with the several closest frames assigned to it.
Observed from the first hierarchy, the model puts a variety of vehicle-related frames into a single node. These frames are then refined into frames about cross-country vehicles and frames about vehicle-repairing. The frames on vehicle-repairing are further divided into
bike repairing and car wheel repairing. These informative hierarchies learned by our model demonstrates its effectiveness of capturing meaningful
hierarchical patterns in the data as well as exhibits interpretability.

\section{Conclusions}\label{sec:conclude}

We presented a new unsupervised learning framework to combine rich nCRP prior with VAEs. This embeds the data into a latent space with rich hierarchical structure, which has more abstract concepts higher up in the hierarchy, and less abstract concepts lower in the hierarchy. We developed a joint optimization framework for variational updates of both the neural and nCRP parameters. 
We showed an application of our model to video data, wherein, 
our experiments demonstrate that our model outperforms other 
models on two downstream tasks, namely, video classification and video retrieval. Qualitative analysis of our model by visualizing the learned hierarchy shows that our model captures rich interpretable
structure in the data.

\vspace{10pt}

{
\noindent{\bf Acknowledgements. }This work was partly funded by NSF IIS1563887, NSF IIS1447676, ONR N000141410684, and ONR N000141712463. Xiaodan Liang is supported by the Department of Defense under Contract No. FA8702-15-D-0002 with Carnegie Mellon University for the operation of the Software Engineering Institute, a federally funded research and development center.
}

\clearpage
\balance
{\small
		\bibliographystyle{ieee}
		\bibliography{refs}
}

\end{document}